\documentclass[11pt]{article}

\usepackage[preprint]{acl}

\usepackage{times}
\usepackage{latexsym}

\usepackage[T1]{fontenc}

\usepackage[utf8]{inputenc}

\usepackage{microtype}

\usepackage{inconsolata}

\usepackage{graphicx}

\usepackage{amsmath}
\usepackage{array}
\usepackage{enumitem}
\usepackage{booktabs}
\usepackage{adjustbox}
\usepackage{array}
\usepackage{booktabs}
\usepackage{colortbl}
\usepackage{arydshln}
\usepackage{bbding} 
\usepackage{amssymb}
\usepackage{multirow}
\usepackage{xcolor}
\usepackage{subcaption}
\usepackage{float}
\usepackage{xcolor}
\usepackage{pifont}
\newcommand{\cmark}{\ding{51}}
\newcommand{\xmark}{\textcolor{gray!50}{\ding{55}}} 
\usepackage[table]{xcolor}
\usepackage{caption}
\usepackage{cuted}

\title{MobileDreamer: Generative Sketch World Model for GUI Agent}

\author{ 
\textbf{Yilin Cao}\textsuperscript{\textbf{1,2}}\thanks{Equal Contribution.},  
\textbf{Yufeng Zhong}\textsuperscript{\textbf{3}}\footnotemark[1],  
\textbf{Zhixiong Zeng}\textsuperscript{\textbf{3}}\footnotemark[2],  
\textbf{Siran Dai}\textsuperscript{\textbf{4,5}},  
\textbf{Liming Zheng}\textsuperscript{\textbf{3}}, \\ 
\textbf{Jing Huang}\textsuperscript{\textbf{3}}, 
\textbf{Haibo Qiu}\textsuperscript{\textbf{3}},  
\textbf{Peng Shi}\textsuperscript{\textbf{3}}, 
\textbf{Wenji Mao}\textsuperscript{\textbf{1,2}}\thanks{Corresponding Author.} \\ 
\textsuperscript{1}State Key Laboratory of Multimodal Artificial Intelligence Systems,\\ 
Institute of Automation, Chinese Academy of Sciences \\ 
\textsuperscript{2}School of Artificial Intelligence, University of Chinese Academy of Sciences \\ 
\textsuperscript{3}Meituan \\ 
\textsuperscript{4}Institute of Information Engineering, Chinese Academy of Sciences \\ 
\textsuperscript{5}School of Cyber Security, University of Chinese Academy of Sciences \\ 
\texttt{wenji.mao@ia.ac.cn, zengzhixiong@meituan.com} 
}

\begin{document}

\maketitle

\begin{abstract}

Mobile GUI agents have shown strong potential in real-world automation and practical applications. 
However, most existing agents remain reactive, making decisions mainly from the current screen, which limits their performance on long-horizon tasks.
To address this limitation, a natural approach is to use the forecasting ability of world models to support better decision-making. 
This is challenging because existing world models often lack spatial awareness when predicting post-action states and are too expensive for practical deployment.
In this paper, we propose MobileDreamer, an efficient world-model-based lookahead framework to equip GUI agents based on the future imagination provided by the world model.
It consists of a textual sketch world model and rollout imagination for GUI agent. Textual sketch world model forecasts post-action states through a learning process to transform digital images into key task-related sketches, and designs a novel order-invariant learning strategy to preserve the spatial information of GUI elements.
The rollout imagination strategy for the GUI agent optimizes the action-selection process by leveraging the prediction capability of the world model.
Empirical results on Android World demonstrate that MobileDreamer consistently improves the task success rate across multiple base models, achieving an average performance gain of $5.3\%$.
World model evaluations further verify that our textual sketch modeling accurately forecasts key GUI elements.

\end{abstract}

\begin{figure*}[t]
  \centering
  \includegraphics[width=0.9\textwidth]{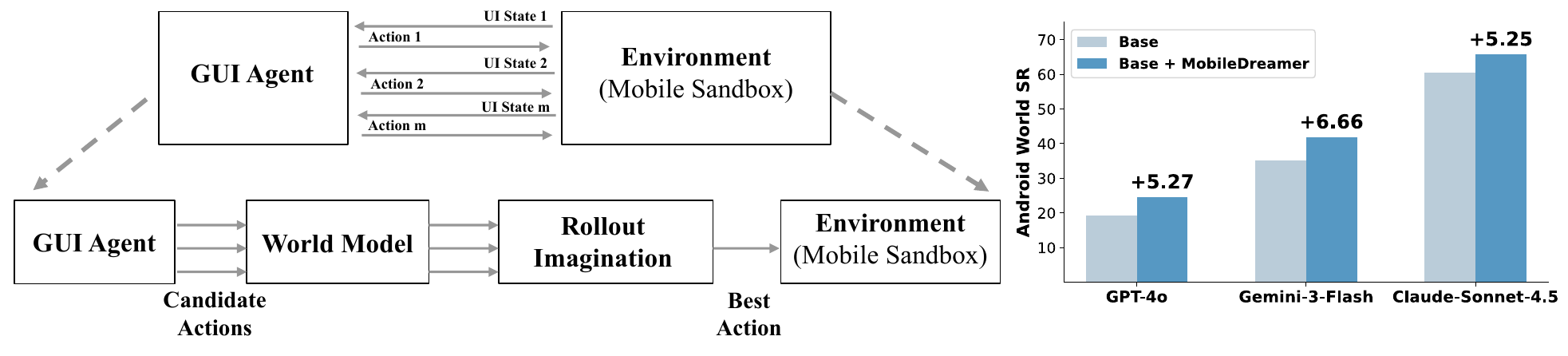}
  \caption{MobileDreamer trains a world model to predict future GUI states and performs rollout imagination to select actions from candidates. It consistently improves multiple LLM backbones on Android World.}
  \label{fig:method1}
\end{figure*}

\section{Introduction}

Mobile GUI agents~\cite{sun2025genesis, sun2025gui, lu2025ui, liu2025infigui} aim to autonomously operate mobile devices by perceiving screen content and executing actions such as clicking, typing, and scrolling to accomplish user-specified tasks. Existing mobile GUI agents have demonstrated significant potential across diverse domains, including web automation~\cite{ye2025mobile, liu2025llm}, mobile app testing~\cite{yoon2024intent, yu2023vision}, and enterprise workflow optimization~\cite{lin2025inframind}.

Despite rapid progress, most mobile GUI agents remain predominantly reactive, making decisions based on the current screen and recent history without anticipating downstream interface changes. In contrast, humans operate on a predictive feedback loop: they mentally simulate the consequences of an action before execution, and subsequently calibrate their future planning by comparing the actual interface response with their internal prediction. Consequently, in complex long-horizon tasks, GUI agents often struggle with effective planning and make short-sighted decisions that lead to inefficient steps or task failure.

Recent progress in world models offers a promising direction to bridge this gap for mobile GUI agents. By predicting post-action states, a world model acts as an internal simulator that helps agents foresee interface evolution and make more forward-looking decisions in long-horizon tasks.
 
Several early studies have attempted to integrate text-based world models with GUI agents~\cite{gu2024your, chae2024web, li2025mobileworldbench}. However, text-based state prediction lacks spatial information, which is essential for accurate action planning in GUI environments. A recent work uses image-based world models~\cite{luo2025vimo} to predict the next GUI screenshot. While this approach captures spatial information, it is computationally expensive and includes task-irrelevant visual details, which limits its practicality for multi-step planning.

Moreover, how to effectively use predicted future states for action selection remains underexplored. Existing GUI agents mainly rely on single-step greedy selection~\cite{sun2025genesis, sun2025gui, li2025mobileworldbench}. This simple strategy restricts the agent to considering only one step of possible outcomes. Lacking the long-term vision for multi-step proactive behavior, GUI agents may struggle to make feasible future decisions.

To address these challenges, we propose \textbf{MobileDreamer}, an efficient sketch world model-based lookahead framework that equips GUI agents with the ability to predict long-term outcomes before taking an action. It consists of two key components: \textbf{1) Textual Sketch World Model (TSWM)}, which predicts task-relevant information in future states using a structured text formulation. To improve the visual fidelity of textual sketches, we introduce an order-invariant learning objective that optimizes element matching with an IoU-aware cost, making training robust to changes in element order and small position shifts. \textbf{2) Rollout Imagination for GUI Agent}, which recursively feeds predicted trajectories back to the GUI agent as a tree-of-prediction, providing forecast evidence to improve the action-selection process.

Empirically, we validate the proposed TSWM on a future-state forecasting task. The results show strong performance on future state forecasting. We further evaluate MobileDreamer on the Android World benchmark \cite{ICLR2025_AndroidWorld}. 
MobileDreamer significantly outperforms open/closed-source models in its ability to predict future states, and achieves significant improvements over several baseline GUI agents, demonstrating the effectiveness of our framework design as illustrated in Figure~\ref{fig:method1}.

The main contributions of our work are summarized as follows:

\begin{itemize}
\item To efficiently represent GUI states while preserving essential spatial information, we propose a textual sketch world model representation in a lightweight and structured textual format, and develop order-invariant learning to forecast its post-action states.
\item We further design a rollout imagination strategy for GUI agent by devising a multi-step lookahead mechanism with tree-of-prediction, which recursively feeds predicted future trajectories to optimize the proactive action-selection process.
\item Experimental results on Android World show that our framework significantly improves GUI agent task success rates. Experiments on world model evaluation further demonstrate the effectiveness of textual sketch world modeling in accurately forecasting key GUI elements.
\end{itemize}

\section{Related Work}
\label{sec:work}
\subsection{World Models}
World models~\cite{ding2024understanding} are learning-based systems that predict environment dynamics to enable intelligent agents to anticipate future states for better decision-making. They have been extensively applied in embodied AI~\cite{cen2025worldvla, yao2025navmorph}, autonomous driving~\cite{wang2025adawm, chen2025drivinggpt}, and game~\cite{hafner2025training, zhou2024minedreamer}, demonstrating strong predictive capabilities.
In the GUI agent domain, early explorations focus on text-based world models that predict interface changes through natural language descriptions. WEBDREAMER~\cite{gu2024your} simulates action outcomes with natural language and uses LLM-based scoring for action selection. WMA~\cite{chae2024web} predicts important state changes through transition-focused abstraction. MobileWorld~\cite{li2025mobileworldbench} represents transitions as natural language triplets. While computationally efficient, these text-based approaches lose critical spatial information such as component positions and bounding boxes.
Recently, ViMo~\cite{luo2025vimo} pioneered image-based world models by predicting next-frame GUI screenshots. However, pixel-level prediction incurs high computational costs and struggles with fine-grained detail reconstruction. Moreover, existing approaches lack effective mechanisms for multi-step lookahead, typically relying on single-step greedy selection.

\subsection{GUI Agents}
Early GUI agents rely on structured data such as HTML or accessibility trees~\cite{deng2023mind2web, gur2023real, lai2024autowebglm, he2024webvoyager, yang2023set}. 
With the advancement of MLLMs, vision-based methods~\cite{Hong_2024_CVPR, zhang2024you, zhang2024android, huang2025scaletrack} have emerged, demonstrating superior generalization by directly processing screenshots. 
UGround~\cite{gou2025navigating} improves GUI agents with universal visual grounding for locating target elements on the current screen.
Building on these foundations, recent works~\cite{qin2025ui, yan2025stepguitechnicalreport, zeng2025uitron} achieve state-of-the-art performance through further innovations in training strategies. 
UI-TARS~\cite{qin2025ui} introduces a native agent approach with long-term memory and reflection mechanisms for iterative improvement.
Step-GUI~\cite{yan2025stepguitechnicalreport} proposes a self-evolving training pipeline that generates high-quality data through trajectory-level calibration.
In parallel, research has also explored cross-domain synergies, with works such as~\cite{han2025guirobotron, yang2025omniactor} integrating GUI agents with robotics and other modalities.
In the mobile agent domain~\cite{sun2025genesis, sun2025gui}, researchers have explored various approaches to enhance agent capabilities. Data-centric methods focus on scaling training data through synthetic trajectories~\cite{sun2025genesis} or exploration videos~\cite{sun2025gui}. 
Despite these advances, existing mobile agents remain largely reactive, making decisions based on current observations without anticipating future consequences. This limitation could potentially be addressed through world models.

\section{Method}
\label{sec:method}

\begin{figure*}[h]
  \includegraphics[width=\linewidth]{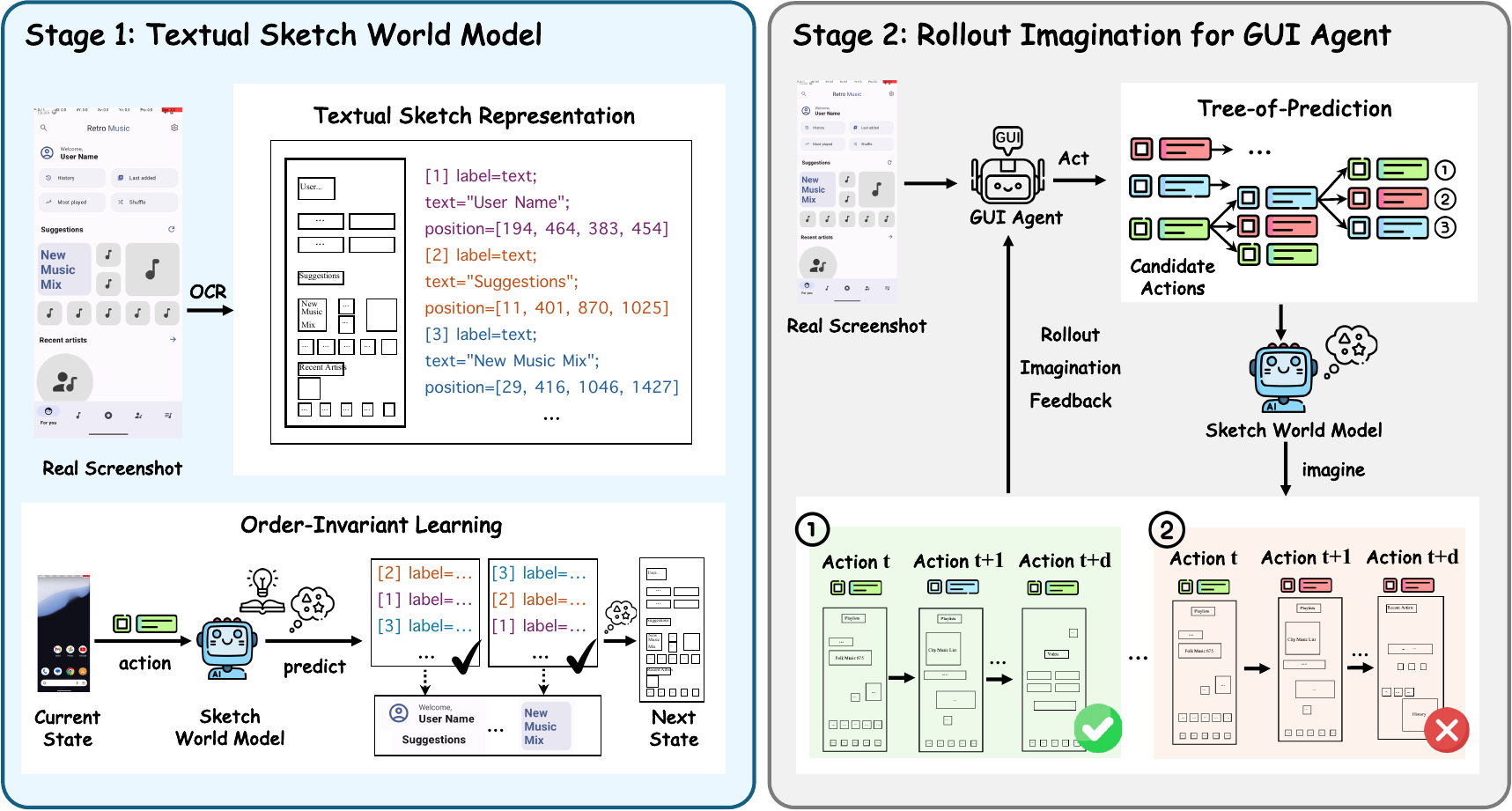}
  \caption {Overview of the MobileDreamer framework. In stage 1, we build a textual sketch world model to predict future states through order-invariant learning. In stage 2, we design a rollout imagination strategy that recursively feeds the tree-of-prediction back to the GUI agent to optimize the action-selection process.}
  \label{fig:method}
\end{figure*}

In this section, we formally present the proposed \textbf{MobileDreamer} framework, which aims to equip GUI agents with the ability to predict long-term outcomes before taking an action. As shown in Figure~\ref{fig:method}, it consists of two stages. In the first stage, we build a Textual Sketch World Model (TSWM), an action-conditioned world model that forecasts post-action GUI states in a lightweight, structured textual format. In the second stage, we propose a rollout imagination strategy for GUI agents, which leverages the TSWM to unroll a multi-step tree-of-prediction over candidate action paths. This look-ahead mechanism allows the agent to select the optimal action based on its simulated downstream utility.

\subsection{Textual Sketch World Model}
\label{sec:wm}

While prior work has attempted to use world-model prediction for action selection \cite{gu2024your, chae2024web, li2025mobileworldbench, luo2025vimo}, relying on diffusion models to reconstruct GUI pages adds substantial cost and introduces prediction-irrelevant details, such as textures and decorations. These details can distract the agent and harm decision making. To address this issue, we propose a Textual Sketch World Model (TSWM) that predicts only the minimal information needed for GUI-agent planning.

\paragraph{Textual Sketch Representation}
We first use PaddleOCR-VL \cite{cui2025paddleocr} to extract key elements from a GUI page. After this process, each icon is represented in a structured format. To balance computational efficiency with a clear description of the layout, we keep three key properties: the label, the text within the icon, and the position on the page. As a result, the current state $s_{t}$ is represented as an element set $\mathcal{E}(s_{t})=\{e_n\}_{n=1}^{N}$, where each element $e_n=(\ell_n,\tau_n,b_n)$ contains label $\ell_n$, text $\tau_n$, and position (bounding box) $b_n$. Below is an example of the extracted data:

\newlength{\sketchboxw}
\setlength{\fboxsep}{5pt} 
\setlength{\sketchboxw}{\dimexpr\columnwidth-2\fboxsep-2\fboxrule\relax}

\noindent\fbox{%
\begin{minipage}{\sketchboxw}
\small\ttfamily\raggedright 
\mbox{label=image;text="Home";bbox=[35,175,347,233]}\par
\mbox{label=text;text="Totals";bbox=[50,343,285,467]}\par
\mbox{label=text;text="INCOME";bbox=[96,519,239,624]}
\end{minipage}%
}

Based on the textual sketch representation, GUI states are encoded as structured text. A straightforward approach is therefore to train the textual sketch world model in a supervised manner. Specifically, the next state $s_{t+1}$ is obtained by executing action $a_t$ on the current state $s_t$, and the world model can be trained via supervised fine-tuning (SFT) with token-level cross-entropy:
\begin{equation} \mathcal{L}_{\text{CE}} = - \sum_{i} \log p_{\theta}\!\left(s_{t+1,i}\mid s_{t+1,<i},\, s_t, a_t\right), \end{equation}
where $s_{t+1,i}$ denotes the $i$-th token in the textual-sketch sequence of the next GUI state, and $\theta$ denotes the trainable parameters.

However, due to the mismatch between the goals of world models and LLMs, we argue that using SFT alone can lead to suboptimal performance. In particular, there are two cases where a prediction can be sufficient for world-model use, yet still incur a large SFT loss:
\begin{itemize}
\item \textbf{Element order change.} The post-action GUI state is naturally a set of elements. Reordering elements (e.g., $[A,B,C]$ vs.\ $[B,C,A]$) should not be treated as incorrect, but token-level supervision is order-sensitive.
\item \textbf{Position micro-shifts.} A small position offset (e.g., a ground-truth bbox \texttt{[100,200,300,400]} vs.\ a predicted bbox \texttt{[101,199,301,399]}) can incur a large token-level loss, even when the geometry is nearly correct ($\mathrm{IoU}\approx 0.98$).
\end{itemize}

\paragraph{Order-Invariant Learning}
Since naive SFT is not well-suited for world-model pretraining, we introduce an element-level matching loss. Intuitively, we treat the TSWM objective as an object detection problem. For each predicted icon, we match it to a ground-truth icon using optimal transport, and then align its properties.

Concretely, given the prediction of TSWM as $\hat{s}_{t+1}$, we denote the predicted element set as $\hat{\mathcal{E}}(\hat{s}_{t+1})=\{\hat{e}_k\}_{k=1}^{K}$, where each predicted element is $\hat{e}_k=(\hat{\ell}_k,\hat{\tau}_k,\hat{b}_k)$. We then compute a one-to-one set matching between predicted and ground-truth elements. Let $\Pi$ denote the set of valid one-to-one matchings; each matching $\pi\in\Pi$ is a set of index pairs $\pi\subseteq\{1,\dots,K\}\times\{1,\dots,N\}$ such that each $k$ and each $n$ appears in at most one pair.

For a predicted--ground-truth pair $(\hat{e}_k,e_n)$, we define a weighted cost function to measure geometry and content mismatches:
\begin{equation}
\begin{aligned}
C(\hat{e}_k,e_n)
&= \lambda_{\mathrm{bbox}} \bigl(1-\mathrm{IoU}(\hat{b}_k,b_n)\bigr) \\
&\quad + \lambda_{\mathrm{label}} \bigl(-\log \hat{p}_k(\ell_n)\bigr) \\
&\quad + \lambda_{\mathrm{text}} \bigl(1-\cos(\phi(\hat{\tau}_k),\phi(\tau_n))\bigr),
\end{aligned}
\end{equation}
where $\hat{p}_k(\ell_n)$ is the predicted probability of label $\ell_n$ for $\hat{e}_k$, and $\phi(\cdot)$ maps an element text string to a fixed-dimensional text embedding. Using this cost function, we compute the optimal transport by minimizing the total pairwise cost:
\begin{equation}
\pi^{\star}
= \operatorname*{arg\,min}_{\pi \in \Pi}
\sum_{(k,n)\in\pi} C(\hat{e}_k, e_n).
\end{equation}

Given the optimal matching $\pi^{\star}$, we compute the set-level matching loss by averaging the pairwise costs over the matched pairs:
\begin{equation}
\mathcal{L}_{\mathrm{match}}
= \frac{1}{|\pi^{\star}|} \sum_{(k,n)\in\pi^{\star}} C(\hat{e}_k,e_n).
\end{equation}

Finally, we optimize the textual sketch world model with:
\begin{equation}
\mathcal{L}
= \mathcal{L}_{\mathrm{match}} + \lambda_{\mathrm{CE}}\,\mathcal{L}_{\mathrm{CE}}.
\end{equation}

This objective encourages the textual sketch world model to preserve element-level structure under action-conditioned transitions, making its forecasts more robust to element reordering and small text or bounding-box perturbations. The resulting predictions provide clearer signals for comparing candidate actions and support more reliable lookahead planning in the Rollout Imagination.

\subsection{Rollout Imagination for GUI Agent}
\label{sec:top}
After obtaining a powerful TSWM, we propose a rollout imagination strategy that utilizes its prediction capability to improve the action-selection process. 

\paragraph{Candidate Action Generation}
Given the task goal $g$, current state $s_t$, and step instruction $u_t$, the GUI agent proposes a set of $M$ candidate actions $\mathcal{A}_t = \{a_t^{(1)}, \dots, a_t^{(M)}\}$. MobileDreamer treats these candidates as alternative branches, which are evaluated through the lookahead process.

For each candidate action $a_t^{(m)}$, the world model predicts the next state:
\begin{equation}
\hat{s}_{t+1}^{(m)} \sim p_{\theta}(s_{t+1} \mid g, s_t, a_t^{(m)}), \quad m \in \{1,\dots,M\}.
\end{equation}
These predictions form a set of future states corresponding to each candidate action, providing additional context to the GUI agent to help it select the best action based on the anticipated future states.

\paragraph{Tree-of-Prediction}

To model longer-term effects, we expand the one-step forecasts into a prediction tree of depth $d$, recursively proposing follow-up candidate actions based on predicted states and forecasting their future states. After the first prediction step, MobileDreamer feeds the predicted state $\hat{\mathcal{S}}_{t+1}$ and the candidate action $\mathcal{A}_t$ back to the GUI agent, which uses the predicted trajectory to select the next predicted actions $\hat{\mathcal{A}}_{t+1}$. Here, $\mathcal{A}_t=\{a_t^{(1)},\dots,a_t^{(M)}\}$ contains a set of $M$ candidate actions, where $M$ serves as a pruning parameter that controls the branching factor of the prediction tree. This process continues until the tree reaches the specified depth $d$. Each root-to-leaf path represents a short predicted trajectory:
\begin{equation}
\mathcal{A}_t \rightarrow \hat{\mathcal{S}}_{t+1} \rightarrow \hat{\mathcal{A}}_{t+1} \rightarrow \hat{\mathcal{S}}_{t+2} \rightarrow \cdots \rightarrow \hat{\mathcal{S}}_{t+d}.
\end{equation}

These trajectories are summarized into a tree-of-prediction with depth $d$, including candidate actions and their predicted post-action states at each step.
The prompt template used to present the prediction tree to the GUI agent is provided in Appendix \ref{subsec:tree_prompt}.

\paragraph{World Model Feedback to GUI Agent}
After the prediction, MobileDreamer feeds the tree-of-prediction back to the GUI agent as additional context for action selection. Specifically, the GUI agent receives the real current state screenshot for grounding, and a textual summary of the tree-of-prediction describing the predicted future states for each candidate action after execution. The GUI agent then selects the best action based on the feedback and executes it in the environment.

Tree-of-prediction allows the GUI agent to explicitly compare alternative actions based on their anticipated states. This is especially crucial in complex GUI navigation tasks that require long-term planning.

\section{Experiments}

\begin{figure}[t]
  \centering
  \begin{subfigure}[t]{0.49\columnwidth}
    \centering
    \includegraphics[width=\linewidth]{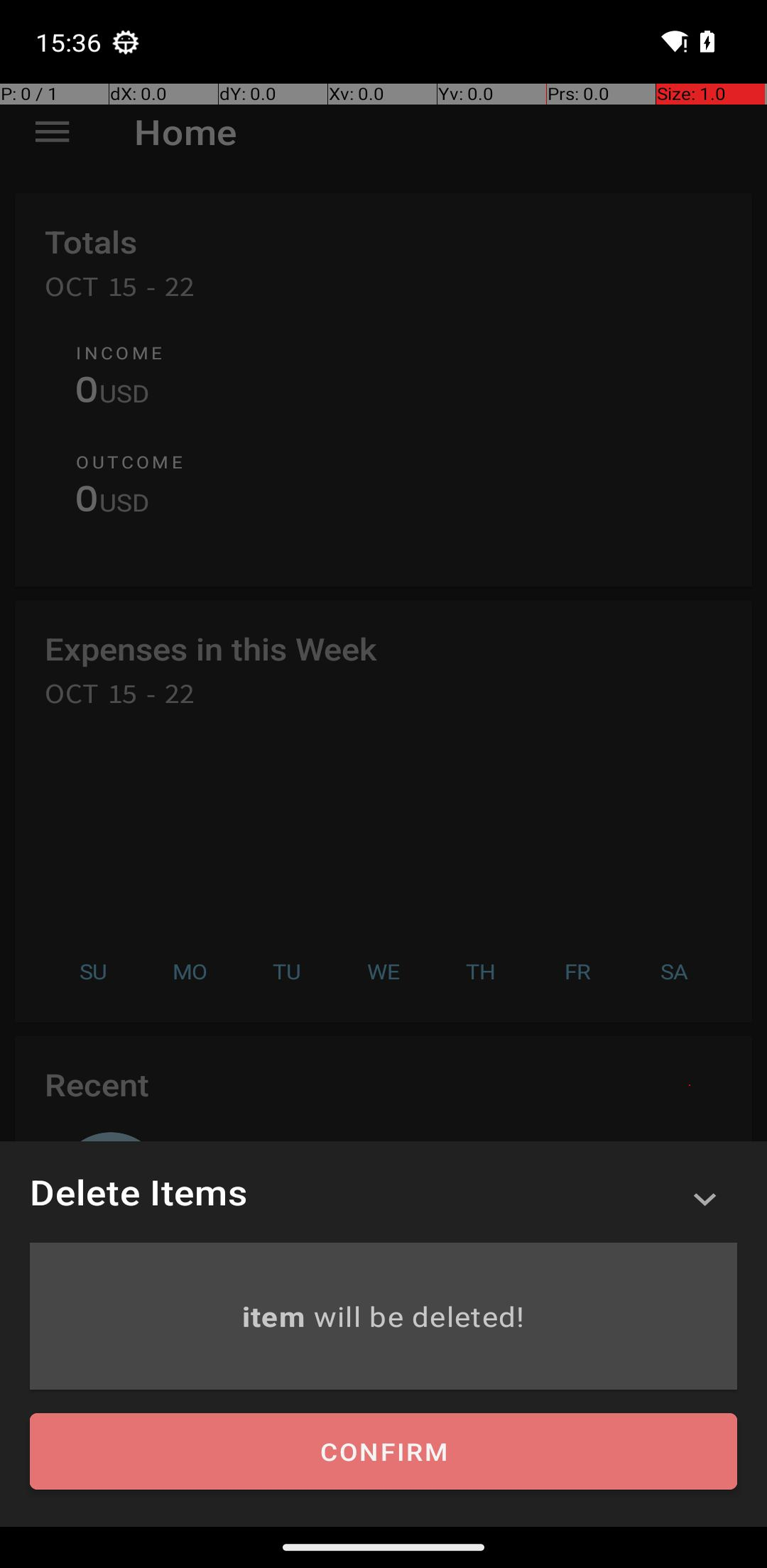}
    \caption{Real screenshot.}
    \label{fig:real-screenshot}
  \end{subfigure}
  \hfill
  \begin{subfigure}[t]{0.49\columnwidth}
    \centering
    \includegraphics[width=\linewidth]{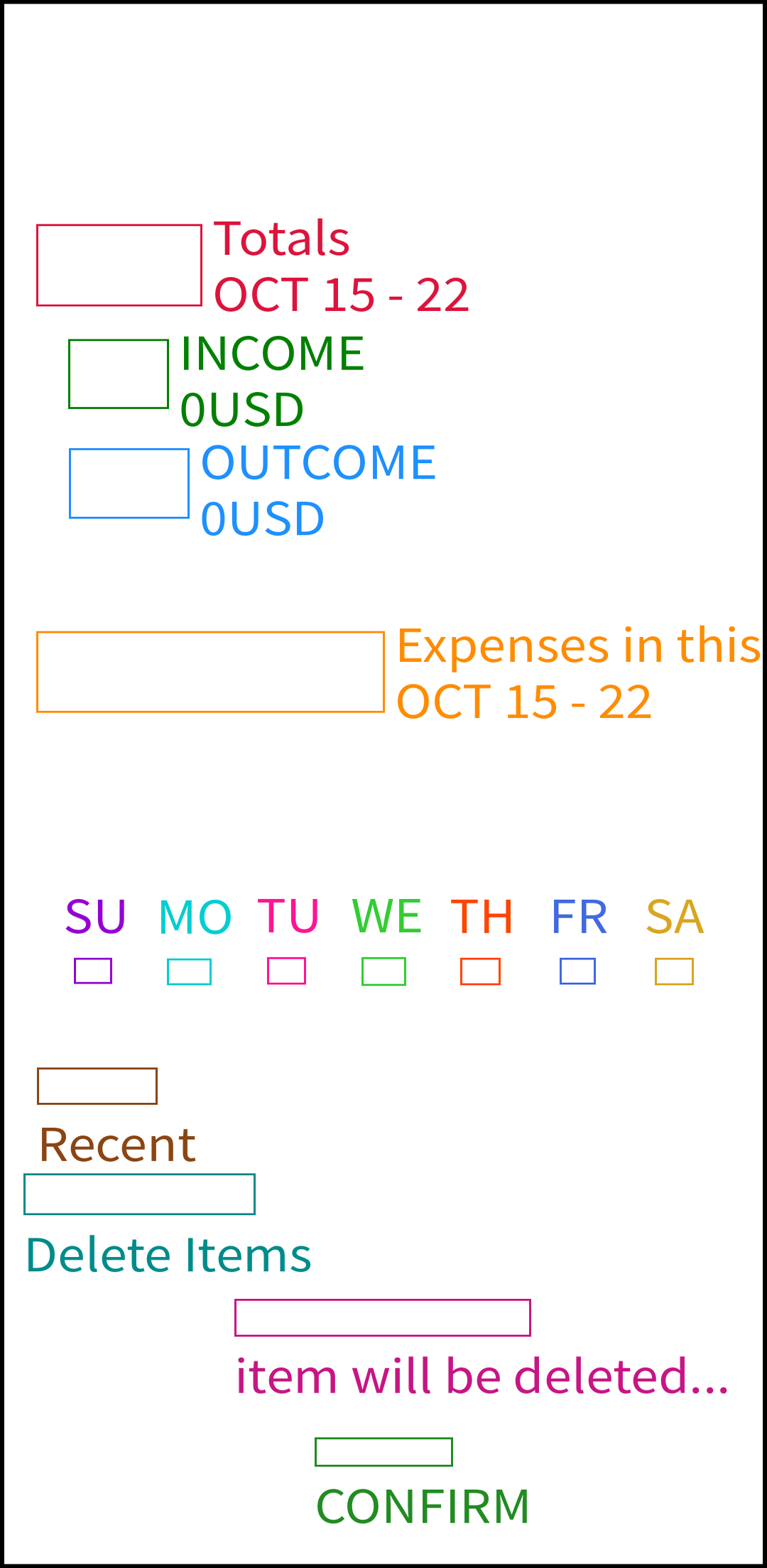}
    \caption{Predicted textual sketch.}
    \label{fig:sketch-pred}
  \end{subfigure}
  \caption{Comparison between the real GUI screenshot and the sketch predicted by our world model. The predicted sketch captures key UI elements with texts and positions. For clarity, we display the element text outside the corresponding element boxes.}
  \label{fig:textual-sketch-compare}
\end{figure}

\begin{table*}[t]
  \centering
  \resizebox{0.95\textwidth}{!}{%
  \begin{tabular}{p{0.35\textwidth}ccccc}
    \toprule
    \textbf{Method} & \textbf{mIoU} & \textbf{Text Similarity} & \textbf{Precision} & \textbf{Recall} & \textbf{F1-Score} \\
    \midrule
    Qwen3-VL-Plus & 0.6016 & 0.8159 & 0.4417 & 0.3893 & 0.4139 \\
    GPT-4.1 & 0.5512 & 0.7967 & 0.4272 & 0.3397 & 0.3784 \\
    Gemini-3-Flash-Preview & 0.4981 & 0.7759 & 0.3792 & 0.1291 & 0.1927 \\
    Claude-Sonnet-4.5 & 0.4668 & 0.7596 & 0.3797 & 0.0980 & 0.1557 \\
    \midrule
    \textcolor{black!50}{Qwen3-8B (baseline)} &
    \textcolor{black!50}{0.6368} & \textcolor{black!50}{0.8224} &
    \textcolor{black!50}{0.4482} & \textcolor{black!50}{0.4223} & \textcolor{black!50}{0.4349} \\
    \textbf{MobileDreamer} & \textbf{0.8564} & \textbf{0.9428} & \textbf{0.8233} & \textbf{0.7071} & \textbf{0.7608} \\
    \bottomrule
  \end{tabular}
}
  \caption{Evaluation of the textual sketch world model on future-state forecasting. We report geometry agreement (mIoU), text agreement (Text Similarity), and element-level matching metrics (Precision/Recall/F1-Score).}
  \label{tab:worldmodel-eval}
\end{table*}

\subsection{Experiments on World Model}

\paragraph{Dataset}
To train and evaluate the textual sketch world model, we construct a dataset of state transition samples from real mobile GUI trajectories.
Given a trajectory, for each step $t$ we collect the pre-action screenshot state $s_t$, the executed action $a_t$ and the post-action screenshot state $s_{t+1}$.
Each resulting sample is an action-conditioned instance $(s_t, a_t, s_{t+1})$.

To convert raw screenshots into structured textual sketches, we annotate each screenshot using PaddleOCR-VL~\cite{cui2025paddleocr} and represent the GUI state as a set of UI elements, each associated with its label, recognized text, and position (bounding box). This conversion enables the textual sketch world model to learn how the GUI state transitions after executing an action.

To ensure diversity and realism, we construct samples from two sources.
First, we leverage the open-source English trajectories from Android Control~\cite{li2024on}, yielding $61.7$k samples for training.
Second, we manually annotate an additional $50.5$k samples.
For evaluation, we reserve $5\%$ trajectories from the manually annotated set as the test split, ensuring there is no trajectory overlap with the training data.
After converting trajectories into textual sketch state representations and splitting them into step-level transitions, the resulting dataset contains $110.1$k training samples and $2.1$k test samples.

\paragraph{Baselines}
To compare different LLM predictors with our sketch world model, we evaluate several prompt-based LLMs, including Qwen3-VL-Plus~\cite{Qwen3-VL}, GPT-4.1~\cite{openai2025gpt41}, Gemini-3-Flash-Preview~\cite{google2025gemini3flash}, and Claude-Sonnet-4.5~\cite{anthropic2025claude45system}. We also use Qwen3-8B~\cite{yang2025qwen3technicalreport} as the backbone of our textual sketch world model, and report results with and without the proposed order-invariant learning objective.

\paragraph{Evaluation Metrics}
We evaluate the performance of the textual sketch world model in forecasting the next state at the element level using mIoU, text similarity, precision, recall, and F1-score. For each test sample, we extract the post-action state from both the ground truth and the prediction, parse the UI elements, and optionally deduplicate the predicted elements based on text. A one-to-one greedy matching is then performed between the predicted and ground-truth elements based on a relaxed matching criterion: an element is considered a match if either their position have an intersection over union (IoU) above a threshold $\theta_{\mathrm{iou}}$ or their texts have a normalized edit distance below a threshold $\theta_{\mathrm{txt}}$. We set $\theta_{\mathrm{iou}}=0.7$ and $\theta_{\mathrm{txt}}=0.3$ in our evaluation. The detailed evaluation metrics is provided in Appendix \ref{subsec:Evaluation Metrics of World Model}.

\paragraph{Model Settings}
We build the textual sketch world model on Qwen3-8B \cite{yang2025qwen3technicalreport}. All models are fine-tuned with LoRA \cite{hu2022lora} on clusters of NVIDIA H100 80GB GPUs. We use a learning rate of $1\times10^{-4}$ and train for 5 epochs. We employ DeepSpeed ZERO3-style data parallelism.


\paragraph{Main Results and Analysis}
Figure~\ref{fig:textual-sketch-compare} compares the predicted sketch state with the ground-truth sketch derived from the screenshot, showing that our sketch world model can accurately recover key UI elements and their spatial layout.
Table~\ref{tab:worldmodel-eval} reports future-state forecasting quality at the element level. Our sketch world model substantially improves both spatial and textual accuracy over prompt-based LLM predictors and the Qwen3-8B~\cite{yang2025qwen3technicalreport} baseline.
These results indicate that our sketch world model can reliably predict key post-action elements, providing higher-quality predicted states for tree-of-prediction planning.

\subsection{Experiments on mobile GUI Agent}

\paragraph{Dataset}
To further demonstrate the effectiveness of MobileDreamer in realistic mobile GUI navigation scenarios, we evaluate MobileDreamer on Android World dataset \cite{ICLR2025_AndroidWorld}. 
Android World is an online benchmark that executes tasks in Android emulators and supports automated evaluation.
It contains 116 distinct navigation tasks across 20 real-world Android apps.
Each task provides standardized initialization, success checking, and teardown procedures, which enables reproducible rollouts and state-based verification.
Task instances are generated by sampling task parameters, which yields diverse inputs and reduces reliance on fixed interaction scripts.


\paragraph{Online Evaluation}

We measure the performance using the task Success Rate (SR). A run is successful if the final device state satisfies the task requirement, verified by the benchmark checker without modifying app source code.

\paragraph{Baselines}

We compare MobileDreamer with two categories of baselines on Android World, following the same environment setup, task splits, and evaluation settings.

\begin{itemize}
  \item \textbf{Baselines with World Model.}
  We compare against previous methods that incorporate a world model for planning, including ViMo's image-based world model~\cite{luo2025vimo} and MobileWorld's text-based world model~\cite{li2025mobileworldbench}.

  \item \textbf{Basic LLM Backbones.}
  To isolate the benefit of MobileDreamer, we also evaluate several backbone models as direct GUI agents, including Qwen3-VL-Plus~\cite{Qwen3-VL}, GPT-4.1~\cite{openai2025gpt41}, GPT-4o~\cite{hurst2024gpt}, Gemini-3-Flash~\cite{google2025gemini3flash}, and Claude-Sonnet-4.5~\cite{anthropic2025claude45system}. For each backbone, we report results with and without MobileDreamer under the same prompting and execution setup.
\end{itemize}


\paragraph{Main Results and Analysis}

\begin{table}[t]
  \centering
  \begin{tabular}{cc}
    \toprule
     \textbf{Method} &
    \begin{tabular}[c]{@{}c@{}}
      \textbf{Android World(SR)} \\
    \end{tabular} \\
    \midrule
    \multicolumn{2}{l}{\textbf{\textit{\textcolor{black!65}{Previous World Model Method}}}} \\
    \cdashline{1-2}[2pt/2pt]
     ViMO$^{\dagger}$ & 40.95 \\
    MobileWorld$^{\ddagger}$ & 54.30 \\   
 \midrule
    \multicolumn{2}{l}{\textbf{\textit{\textcolor{black!65}{Our MobileDreamer}}}} \\
    \cdashline{1-2}[2pt/2pt]
   
    Qwen3-VL-Plus & 9.57 \\
       +MobileDreamer & 13.91 (+4.34) \\
       \cdashline{1-2}[2pt/2pt]
    GPT-4.1 & 17.11 \\
    +MobileDreamer & 22.12 (+5.01) \\
    \cdashline{1-2}[2pt/2pt]
     GPT-4o & 19.29 \\
    +MobileDreamer & 24.56 (+5.27) \\
    \cdashline{1-2}[2pt/2pt]
     Gemini-3-Flash & 35.24 \\
    +MobileDreamer & 41.90 (+6.66) \\
    \cdashline{1-2}[2pt/2pt]
     Claude-Sonnet-4.5 & 60.53 \\
    +MobileDreamer & \textbf{65.78 (+5.25)} \\
    \bottomrule
  \end{tabular}
  \caption{Comparison of our MobileDreamer and baselines on Android World. SR represents the task success rate. $^{\dagger}$Results are reported from ViMo \cite{luo2025vimo}. $^{\ddagger}$Results are reported from MobileWorld \cite{li2025mobileworldbench}. For MobileDreamer, we use the tree-of-prediction with depth $d{=}2$ and candidate actions $M{=}3$.}
  \label{tab:androidworld-sr}
\end{table}

\begin{table}[t]
  \centering
  \renewcommand{\arraystretch}{1.15}

  \begin{tabular*}{\columnwidth}{@{\extracolsep{\fill}}cccc}
    \toprule
    \textbf{Sketch} & \textbf{Match} & \textbf{Tree-of-Prediction} & \textbf{SR}$\uparrow$ \\
    \midrule
    \xmark & \xmark & \xmark & 60.53 \\
    \cmark & \xmark & \xmark & 61.94 \\
    \cmark & \cmark & \xmark & 63.16 \\
    \cmark & \cmark & \cmark & 65.78 \\
    \bottomrule
  \end{tabular*}
  \caption{Ablation on key components of MobileDreamer on Android World. \textbf{Sketch} denotes SFT on the textual sketch world model. \textbf{Match} denotes Order-Invariant Learning. \textbf{SR} denotes the task success rate on Android World.}
  \label{tab:ablation}
\end{table}

\begin{table*}[t]
  \centering
  \resizebox{0.85\textwidth}{!}{%
  \begin{tabular}{p{0.30\textwidth}ccccc}
    \toprule
    \textbf{Method} & \textbf{mIoU} & \textbf{Text Similarity} & \textbf{Precision} & \textbf{Recall} & \textbf{F1-Score} \\
    \midrule
    \textbf{MobileDreamer} & \textbf{0.8564} & \textbf{0.9428} & \textbf{0.8233} & \textbf{0.7071} & \textbf{0.7608} \\
    \hspace{0.6em}w/o Order-Invariant Learning & 0.8281 & 0.9320 & 0.8039 & 0.6905 & 0.7429 \\
    \bottomrule
  \end{tabular}%
  }
  \caption{Ablation study on the textual sketch world model. We compare our MobileDreamer with its SFT-only variant that removes the proposed order-invariant learning objective.}
  \label{tab:worldmodel-ablation}
\end{table*}

\begin{table}[t]
  \centering
  \resizebox{0.8\columnwidth}{!}{%
  \begin{tabular}{ccc}
    \toprule
    \textbf{Depth(s)} & \textbf{Action(s)} &
    \begin{tabular}[c]{@{}c@{}}
      \textbf{Android World(SR)} \\
    \end{tabular} \\
    \midrule
    1 & 1 & 63.16 \\
    2 & 1 & \textbf{64.28} \\
    3 & 1 & 63.96 \\    
    \cdashline{1-3}[2pt/2pt]
    2 & 1 & 64.28 \\
    2 & 2 & 64.86 \\
    2 & 3 & \textbf{65.78} \\
    2 & 4 & 65.48 \\    
    2 & 5 & 65.17 \\
    \bottomrule
  \end{tabular}%
  }
  \caption{Ablation on candidate actions and depths of Tree-of-Prediction on Claude-Sonnet-4.5. We vary prediction depth and the number of candidate actions per node. SR represents the task success rate on Android World.}
  \label{tab:ablation-top}
\end{table}

Table~\ref{tab:androidworld-sr} reports success rates on Android World.
MobileDreamer consistently improves task success across all evaluated backbones by introducing sketch world-model-based predicted trajectories into action selection.
It outperforms basic LLM agents and also surpasses prior baselines that incorporate text-based or image-based world models. These results demonstrate the effectiveness of sketch world modeling and the recursive tree-of-prediction feedback for improving GUI action-selection.
Overall, the results validate that our framework enables more effective long-horizon action-selection than reactive or single-step methods. 
Additional results detailed in Appendix \ref{sec:m3a_style_results} further demonstrates the generalization of our approach.

\begin{figure}[t]
  \centering
  \includegraphics[width=\columnwidth]{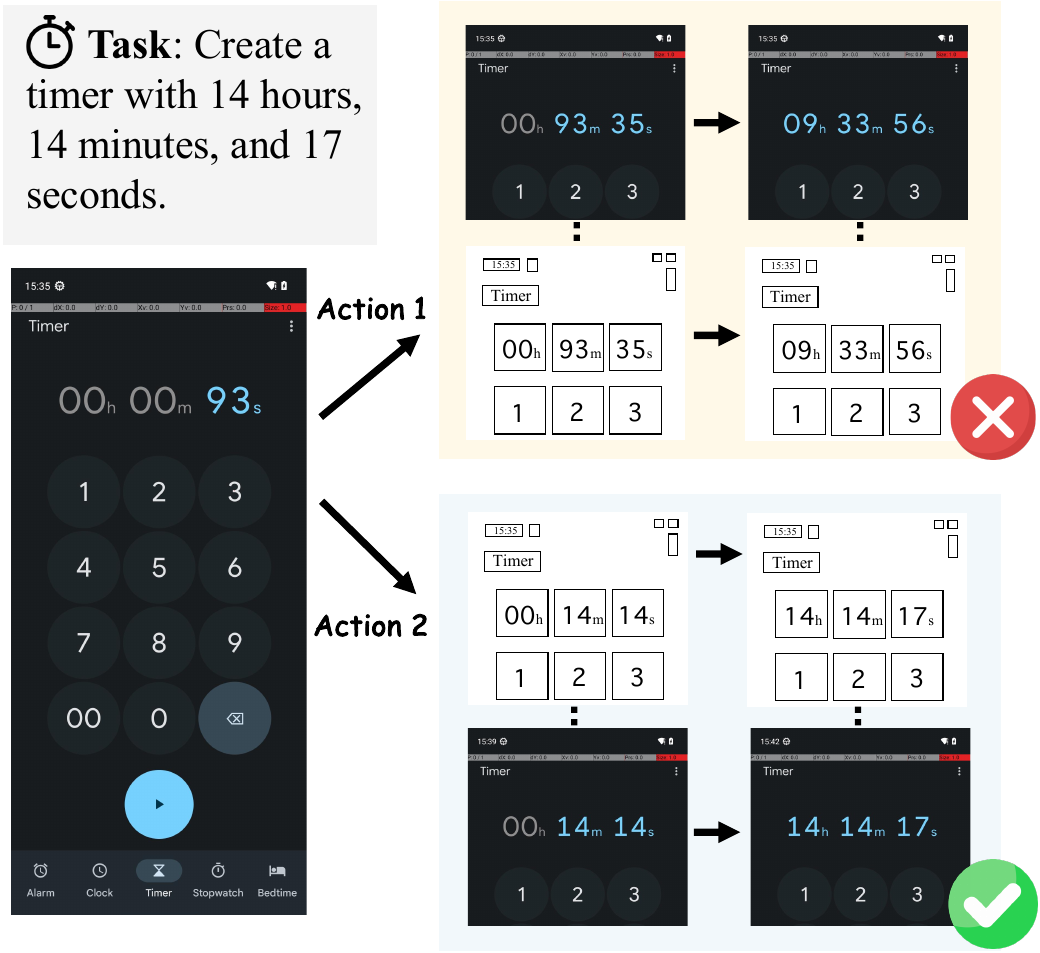}
  \caption{Case study of MobileDreamer on a timer-creation task.}
  \label{fig:case}
\end{figure}

\subsection{Ablation Study}
\label{sec:ablation}

\paragraph{Ablation Study on Key Components}
Table~\ref{tab:ablation} ablates key components in MobileDreamer under Claude-Sonnet-4.5.
Without the sketch world model and tree-of-prediction, the agent degenerates to a reactive baseline.
Adding the sketch world model improves performance by providing action-conditioned predicted states in a structured element layout.
Order-invariant learning further boosts results by producing more robust next-state sketches that better preserve key spatial cues.
Finally, enabling tree-of-prediction achieves the best performance, showing that recursively using predicted trajectories is important for long-horizon action selection.

\paragraph{Ablation Study on Textual Sketch World Model}
Table~\ref{tab:worldmodel-ablation} shows that SFT substantially improves next-state forecasting over the base model, validating that the textual sketch representation provides an effective and learnable target for element-level state transitions.
MobileDreamer further improves over SFT, showing that the order-invariant learning we designed strengthens sketch prediction by making it robust to element reordering and small spatial position noises. Evaluation on the robust performance of TSWM is detailed in Appendix \ref{sec:ocr_reliability}.

\paragraph{Ablation Study on Rollout Imagination Strategy}
Table~\ref{tab:ablation-top} studies tree-of-prediction settings by varying prediction depth and the number of candidate actions per node.
Increasing depth from one step to two steps improves performance, since short-horizon rollouts provide additional future evidence for action selection.
Further increasing depth brings diminishing returns and can be less stable, as deeper recursive feedback introduces more noisy and task-irrelevant predicted information.
Similarly, increasing the number of candidate actions improves performance up to a moderate branching factor, but saturates when too many actions are included.
Overall, a moderate depth and branching factor achieve the best trade-off between useful future evidence and prediction noise.
Furthermore, to verify the practical deployment feasibility of MobileDreamer, supplementary empirical analyses regarding agent step efficiency and framework inference cost are compiled in Appendix \ref{sec:step_efficiency} and \ref{sec:inference_cost}.

\subsection{Case Study}

The case study in Figure~\ref{fig:case} illustrates that the sketch world model provides useful evidence by forecasting future states, which can help the GUI agent select the action most aligned with the task goal. Given two candidate actions at each step, the textual sketch world model forecasts action-conditioned future states and rolls out the tree-of-prediction with depth $2$. The predicted trajectories are recursively fed back to the GUI agent and guides it to select the best action.

\section{Conclusion}

In this paper, we propose MobileDreamer, an efficient sketch world model-based lookahead framework for mobile GUI agents. Our framework builds a lightweight textual sketch world model with order-invariant learning and further designs a rollout imagination strategy that recursively feeds predicted trajectories back to optimize the proactive action selection process. Experimental results show that textual sketch world model enables more effective post-action state predictions for GUI agents on Android World.

\section*{Limitations}
Despite the improvements, MobileDreamer relies on a textual sketch representation extracted by OCR and UI element parsing. Prediction quality may be less stable on screens with heavy visual noise, uncommon fonts, or icon-dominant layouts where reliable text cues are limited. While the sketch representation captures spatial layout for most tasks, it can be less informative for cases that require fine-grained visual signals, such as precise text reading or distinguishing subtle UI appearance differences. In future work, we will improve sketch extraction and incorporate complementary visual cues to better support such scenarios.

\section*{Ethical Considerations}
All experiments are performed in virtual environments without accessing real user systems. We use only publicly available datasets and synthetic data, with no personal information involved.



\section*{Acknowledgments}
This work is supported in part by the National Natural Science Foundation of China under Grant \#72293575.

\bibliography{arxiv_20260831}

\clearpage

\appendix

\section{OCR Reliability Analysis}
\label{sec:ocr_reliability}

The proposed textual sketch representation is constructed from OCR-based UI element extraction.
Therefore, OCR errors such as misread text, hallucinated elements, or missed elements may affect the quality of the extracted sketches.
To quantify the error in the OCR process, we conduct a human evaluation on 300 randomly sampled screenshots with three annotators. These annotators were volunteers and selected based on their proficiency in English, ensuring they have sufficient daily communication skills.

For each extracted UI element, annotators label it as \textit{Correct}, \textit{Misread}, or \textit{Hallucinated} by comparing the extracted text and location with the screenshot.
In addition, annotators mark \textit{missed elements}.
These are visible UI elements that fall within the scope of our sketch schema.
They are elements the extractor is expected to output, but they are absent from the OCR output.
Based on these annotations, we report element-level Precision, Recall, and F1.
The results are shown in Table~\ref{tab:ocr_element_eval}.

We also report three screenshot-level error rates in Table~\ref{tab:ocr_screen_eval}.
Each rate is computed as the fraction of screenshots that contain at least one error of the corresponding type.
These screenshot-level error types are not mutually exclusive, so the rates do not sum to 100\%.

The results show that the extracted sketches achieve high element-level Precision and F1, suggesting that the OCR-based sketch extraction pipeline is reliable in most cases.
At the same time, the miss screen rate indicates that missing elements remain a limitation of the current framework.
In our setting, the GUI agent still receives the real current screenshot as its primary observation.
MobileDreamer adds predicted future sketches as an additional reference for action selection.
As a result, OCR-related noise may affect the lookahead reference, but the predicted future sketch can still provide useful evidence for choosing the next action.

\section{Additional Results under the M3A-style Agent Pipeline}
\label{sec:m3a_style_results}

To further evaluate MobileDreamer under a stronger agent backbone, we conduct additional experiments with the M3A-style agent pipeline on Android World.
This pipeline includes ReAct-style reasoning and self-reflection to summarize progress and improve subsequent decisions, and is also used by UGround~\cite{gou2025navigating} for reporting Android World results with GPT-4 and GPT-4o.
In our main experiments, we use a more minimal GUI-agent backbone without M3A-style selection or reflection, so the absolute success rates are not directly comparable.
We therefore evaluate MobileDreamer on top of the M3A-style pipeline with GPT-4 and GPT-4o to test whether it still provides consistent gains under a stronger agent backbone.
The results are shown in Table~\ref{tab:m3a_style_results}.

\begin{table*}[t]
  \centering
  \begin{tabular}{lccccc}
    \toprule
    \textbf{Setting} & \textbf{Depth} & \textbf{Actions / Node} & \textbf{Time / Step (s)} & \textbf{Overhead vs. Greedy} & \textbf{SR (\%)} \\
    \midrule
    Greedy Selection & 0 & 1 & 10.2 & 1.00$\times$ & 60.53 \\
    Tree-of-Prediction & 1 & 1 & 12.3 & 1.21$\times$ & 63.16 \\
    Tree-of-Prediction & 2 & 1 & 15.8 & 1.55$\times$ & 64.28 \\
    Tree-of-Prediction & 3 & 1 & 19.5 & 1.91$\times$ & 63.96 \\
    Tree-of-Prediction & 2 & 2 & 18.6 & 1.82$\times$ & 64.86 \\
    Tree-of-Prediction & 2 & 3 & 19.4 & 1.90$\times$ & \textbf{65.78} \\
    Tree-of-Prediction & 2 & 4 & 20.2 & 1.98$\times$ & 65.48 \\
    Tree-of-Prediction & 2 & 5 & 21.0 & 2.06$\times$ & 65.17 \\
    \bottomrule
  \end{tabular}
  \caption{End-to-end mean wall-clock time per decision step and success rate under different rollout settings. World model inference is parallelized across nodes at the same rollout depth. SR denotes task success rate.}
  \label{tab:inference_cost}
\end{table*}

\begin{table}[t]
  \centering
  \resizebox{\columnwidth}{!}{%
  \begin{tabular}{ccc}
    \toprule
    \textbf{Element Precision (\%)} & \textbf{Element Recall (\%)} & \textbf{Element F1 (\%)} \\
    \midrule
    96.8 & 88.7 & 92.6 \\
    \bottomrule
  \end{tabular}%
  }
  \caption{Element-level manual evaluation of PaddleOCR-VL sketches on 300 randomly sampled screenshots with three annotators.}
  \label{tab:ocr_element_eval}
\end{table}

\begin{table}[t]
  \centering
  \resizebox{\columnwidth}{!}{%
  \begin{tabular}{ccc}
    \toprule
    \textbf{Misread Screen Rate (\%)} & \textbf{Hallucination Screen Rate (\%)} & \textbf{Miss Screen Rate (\%)} \\
    \midrule
    12.5 & 5.2 & 23.4 \\
    \bottomrule
  \end{tabular}%
  }
  \caption{Screenshot-level OCR error rates on the same 300 randomly sampled screenshots. Each rate is the fraction of screenshots with at least one error of that type.}
  \label{tab:ocr_screen_eval}
\end{table}

\begin{table}[t]
  \centering
  \resizebox{\columnwidth}{!}{%
  \begin{tabular}{lllc}
    \toprule
    \textbf{Source} & \textbf{Method} & \textbf{Backbone} & \textbf{SR (\%)} \\
    \midrule
    UGround~\cite{gou2025navigating} & UGround & GPT-4 & 31.0 \\
    Ours & M3A baseline & GPT-4 & 26.4 \\
    Ours & MobileDreamer + M3A & GPT-4 & 31.6 (+5.2) \\
    \midrule
    UGround~\cite{gou2025navigating} & UGround & GPT-4o & 44.0 \\
    Ours & M3A baseline & GPT-4o & 39.4 \\
    Ours & MobileDreamer + M3A & GPT-4o & 45.1 (+5.7) \\
    \bottomrule
  \end{tabular}%
  }
  \caption{Additional results under the M3A-style agent pipeline on Android World. UGround numbers are reported from~\citet{gou2025navigating}. Our additional results use the M3A-style agent pipeline. SR denotes task success rate.}
  \label{tab:m3a_style_results}
\end{table}

These results show that MobileDreamer provides consistent gains under a stronger agent pipeline.
On GPT-4, MobileDreamer improves the M3A baseline from 26.4\% to 31.6\%.
On GPT-4o, MobileDreamer improves the M3A baseline from 39.4\% to 45.1\%.
This suggests that MobileDreamer is not merely tied to a minimal baseline agent, but can also improve stronger M3A-style GUI agents through action-conditioned future-state prediction.

\section{Step Efficiency Analysis}
\label{sec:step_efficiency}

To further analyze whether lookahead helps the agent make more efficient decisions, we measure task efficiency on Claude-Sonnet-4.5 by comparing the average number of steps with and without the textual sketch world model.
We focus on tasks that succeed in both settings, which yields 64 overlapping successful tasks.
The results are shown in Table~\ref{tab:step_efficiency}.

\begin{table}[t]
  \centering
  \begin{tabular}{lcc}
    \toprule
    \textbf{Method} & \textbf{Avg. Steps} & \textbf{Reduction} \\
    \midrule
    Baseline & 10.8 & -- \\
    MobileDreamer & 8.7 & 19.4\% \\
    \bottomrule
  \end{tabular}
  \caption{Step efficiency analysis on 64 overlapping successful tasks with Claude-Sonnet-4.5. We compare the average number of steps used by the baseline and MobileDreamer.}
  \label{tab:step_efficiency}
\end{table}

The baseline uses 10.8 steps on average, while MobileDreamer uses 8.7 steps on average, which is 19.4\% fewer steps.
This suggests that lookahead helps the agent stay on track and reduces unnecessary interactions when both methods can complete the task.

\section{Inference Cost Analysis}
\label{sec:inference_cost}

We further analyze the inference cost of Tree-of-Prediction.
We log end-to-end wall-clock time at each decision step and report the mean time per step averaged over all tasks.
We vary the rollout depth and the number of candidate actions per node, and report the relative overhead against greedy selection.
All timings are measured with Claude-Sonnet-4.5.
World model inference is parallelized across nodes at the same rollout depth.
The results are summarized in Table~\ref{tab:inference_cost}.

Overall, Tree-of-Prediction yields consistent SR improvements with additional per-step overhead.
Since the textual sketch world model predicts future states in a compact text sketch space, the rollout overhead remains manageable relative to greedy selection.
The best-performing setting, with depth 2 and 3 actions per node, improves SR from 60.53\% to 65.78\%, with a 1.90$\times$ total per-step latency compared with greedy selection.

\section{Prompt Templates}
\label{sec:prompt_templates}

\subsection{Tree-of-Prediction Prompt Template}
\label{subsec:tree_prompt}

The template of tree-of-prediction in MobileDreamer is shown in Table ~\ref{tab:tree_prompt_templates}.

\begin{table}[t]
\centering
\caption{Prompt templates used in Tree-of-Prediction. The top block is used to feed the prediction tree to the GUI agent for action selection. The bottom block is used by the Actor to propose follow-up candidate actions from a predicted textual-sketch state when depth $>1$.}
\label{tab:tree_prompt_templates}
\setlength{\fboxsep}{4pt}
\fbox{
\begin{minipage}{0.97\linewidth}
\footnotesize
\ttfamily
\raggedright

\textbf{[Reasoner] Action selection with Tree-of-Prediction (depth $d{=}2$).}\par
You are selecting the best action based on World Model prediction tree.\par
\par
Task: \{GOAL\}\par
\par
Current sub-task: \{SUBTASK\}\par
\par
The World Model has built a PREDICTION TREE showing what will happen for each candidate action:\par
\{TREE\_TEXT\}\par
\par
Based on the CURRENT SCREEN (shown in the image) and this prediction tree, select the best action.\par
\par
Consider:\par
1. Immediate effect (1st level prediction)\par
2. Long-term trajectory (2nd level predictions if available)\par
3. How well the predicted states align with what you see on the current screen and the task goal\par
\par
Reply with ONLY:\par
<selection>\par
Action Number: [1-\{N\}]\par
Reason: [Your reasoning]\par
</selection>\par

\hrule

\textbf{[Actor] Follow-up candidate actions from a predicted state (used when depth $>1$).}\par
Based on a PREDICTED screen state (not a real screenshot), perform the next action to progress toward the task.\par
\par
Task: \{GOAL\}\par
Current sub-task: \{SUBTASK\}\par
Previous action taken: \{PARENT\_ACTION\}\par
\par
[Predicted Screen State (Text Description)]\par
\{PREDICTED\_STATE\}\par
\par
Based on this predicted screen state, what would be the best next action?\par
\par
Generate \{K\} different candidate actions in JSON format.\par
Each action should be one of:\par
- Click: \{"action": "click", "coordinate": [x, y]\}\par
- Type: \{"action": "type", "text": "content"\}\par
- Scroll: \{"action": "scroll", "scroll\_direction": "up/down/left/right", "coordinate": [x, y]\}\par
- Wait: \{"action": "wait"\}\par
\par
Reply with \{K\} actions, one per line:\par

\end{minipage}
}
\end{table}

\subsection{GUI Agent Instruction Template}
\label{subsec:agent_prompt}

The template of the GUI agent for task planner is shown in Table \ref{tab:planner_prompt}.

\begin{table*}[t]
\centering
\caption{Planner prompt used for prompt-based GUI-agent baselines in Android World.}
\label{tab:planner_prompt}
\fbox{
\begin{minipage}{0.97\textwidth}
\scriptsize
\ttfamily
\raggedright
You are operating an Android phone as a GUI task planner. You will be given the current screenshot of the phone, together with the user instruction you need to fulfill. \par
Your task is to decide what operations should be performed in the following steps, based on the action history, your previous thoughts, current observation, and most importantly, your knowledge about common GUI tasks and the software you are operating. \par
Then, make the decision about the next sub-task starts from the current screen in natural language, together with a general multi-sub-task, long-horizon plan to accomplish the whole task. You can also provide your analyses and thoughts together with the long-horizon plan. \par
Your next sub-task will be executed by another assistant (i.e., actor), so you do not need to give accurate low-level actions such as clicking an element. Instead, give a general description of the next operation, such as "Go into the Hotel page, the icon on the top-left of the screen may be useful". Describe the characteristics of the element you want to operate. \par
Before you make decisions in each step, check whether the outcome of the last few actions has fulfilled your previous plan and operational intent. Write it down in the <reflection></reflection> section. Do not write your plan or thought here. \par
After making decision, briefly describe what you expect to see on the screen after your next action. \par
If there is information relevant to the task that you need to keep in memory on the current screen, such as a task list to be reported to the user, write it down as a note for further use. Otherwise, do not take notes. \par

Be aware that: \par
\ \ * Your next sub-task may contain several steps. If no action should be done now, use the "wait" operation. \par
\ \ \ \ Do not specify how to execute an action such as "click the xxx icon" or "scroll down", you just need to describe what to do, and the actor will decide how to fulfill your operation. \par
\ \ \ \ For example, use "select the file README.md", "see more items below the current screen" instead of "click/long press on the README.md file to select it", "scroll down to see more items". \par
\ \ * You may need to infer the meaning of items from both their titles and icons. \par
\ \ * The actor may take several steps to fulfill the operation, so if the actor has not finished your last operation, be patient for a few steps. Note that some sub-actions in an operation may not change the screen, such as clicking the input field, so please remind the actor about what it has done in the last step in this case. \par
\ \ \ \ Moreover, the actor may not correctly execute your next operation, or your decision may be incorrect. If you find that the result of an operation does not match your expectations, promptly revise your plan and correct any errors, or let the actor try again. \par
\ \ \ \ NEVER post the exact same content as your previous reply. Add a description of the actor's actions if you need to repeat the last operation. \par
\ \ \ \ You may navigate back to the previous page to withdraw the wrong action or take other actions to correct it. \par
\ \ \ \ If the task is stuck for over 3 steps, report "infeasible" to the actor. You should remind the actor the first time you find the action is stuck. \par
\ \ * Update your long-horizon plan and notes after each execution. \par
\ \ * To open an APP, simply state "Open the [APP name]" in your operation. Do not use search or manual navigation methods such as "Scroll down to open the APP drawer and find the camera icon". \par
\ \ \ \ Available APPs are: \{APP\_LIST\}. \par
\ \ * Pay attention to the implicit meanings of text formats. For example, gray text may indicate that a task is complete in a task list, or default content in an input field. Consider these meanings when fulfilling the user instruction. \par
\ \ * When you want to input text, just remind the actor whether to clear the existing contents. Do not take clearing as an independent operation. Directly give the content you want to input. DO NOT use the keyboard on the screen. \par
\ \ * Ignore system errors (e.g., failed messages) and continue with remaining tasks. \par
\ \ * For scroll action: you are NOT using a natural swipe, which means the "up" direction is to view the content above the current screen, or to swipe your finger down from top to bottom, and so on for other directions. \par
\ \ * Avoid unnecessary steps. Always try to complete the task in the fewest possible steps. \par
\ \ * The user instruction might be inaccurate, so consider how humans understand this instruction. \par
\ \ * If you need to use the content of an image, do not use thumbnails, click to check full content of the image. \par
\ \ * If you think the task is successful or infeasible, respond "Ignore the screenshot, terminate the task with status 'xxx' using the terminate tool" in the next operation part. \par
\ \ * If the user instruction asks you to answer a question, summarize from observations and notes, then respond "Ignore the screenshot, answer with the sentence: 'xxx' using the answer tool" in the next operation part. Follow the user's format requirements strictly. \par
\ \ \ \ Attention: Get full information if the text related to your task is truncated (such as ending with "..." or cut off by the screen bottom). \par

You should reply strictly according to the following format: \par
<reflection>Check the outcome of previous actions here.</reflection> \par
<plan>Your long-horizon plan here.</plan> \par
<thought>Your thought about the next SINGLE step.</thought> \par
<operation>Your next sub-task here.</operation> \par
<expection>The expected screen content after operation.</expection> \par
<note>Write down notes here. Can be empty.</note>
\end{minipage}
}
\end{table*}

\section{Evaluation Metrics of World Model}
\label{subsec:Evaluation Metrics of World Model}

\paragraph{mIoU} mIoU measures the average IoU over all matched element pairs, reflecting the geometric consistency between predicted and ground-truth layouts. 

\paragraph{Text Similarity} Text similarity measures the textual agreement between matched elements, computed from the Levenshtein edit distance as a normalized similarity:
\[
\text{TextSimilarity}(s_1, s_2)=1-\frac{\text{EditDistance}(s_1, s_2)}{\max(|s_1|,|s_2|)}.
\]

\paragraph{Precision}
Precision quantifies the proportion of correctly matched predicted elements among all predicted elements. It is calculated as:
\begin{equation}
P = \frac{\mathrm{TP}}{\mathrm{TP} + \mathrm{FP}},
\end{equation}
where $\mathrm{TP}$ is the number of true positives, i.e., predicted elements that are correctly matched to ground-truth elements, and $\mathrm{FP}$ is the number of false positives, i.e., predicted elements that do not match any ground-truth element.

\paragraph{Recall}
Recall measures the proportion of correctly matched ground-truth elements among all ground-truth elements. It is calculated as:
\begin{equation}
R = \frac{\mathrm{TP}}{\mathrm{TP} + \mathrm{FN}},
\end{equation}
where $\mathrm{FN}$ is the number of false negatives, i.e., ground-truth elements that are not matched by any predicted element.

\paragraph{F1-Score}
The F1-score is the harmonic mean of precision and recall, providing a balanced measure of element-level alignment quality:
\begin{equation}
F1 = \frac{2PR}{P + R}.
\end{equation}
This metric balances precision and recall, making it robust to minor variations in positions (bounding box) and small OCR text noise, due to the flexible matching criterion used in the evaluation.

\end{document}